\documentclass{article}
\usepackage[preprint]{neurips_2026}
\usepackage[utf8]{inputenc}
\usepackage[T1]{fontenc}
\usepackage[colorlinks=true,linkcolor=blue,citecolor=blue,urlcolor=blue,anchorcolor=black]{hyperref}
\usepackage{url}
\usepackage{booktabs}
\usepackage{multirow}
\usepackage{amsfonts}
\usepackage{amsmath}
\usepackage{amssymb}
\usepackage{nicefrac}
\usepackage{microtype}
\usepackage{graphicx}
\usepackage{xcolor}
\usepackage{subcaption}
\usepackage{float}
\usepackage{placeins}
\usepackage{caption}

\makeatletter
\renewcommand{\paragraph}{%
  \@startsection{paragraph}{4}{\z@}%
                {0.6ex \@plus 0.3ex \@minus 0.1ex}%
                {-1em}%
                {\normalsize\bfseries\color{black}}%
}
\makeatother

\newcommand{\KL}{\mathrm{KL}}

\title{When Mean CE Fails: Median CE Can Better Track Language Model Quality}

\author{%
  Hao Guo$^{1}$ \qquad
  Simon Dennis$^{1,2}$ \qquad
  Rivaan Patil$^{1,3}$ \qquad
  Kevin Shabahang$^{1}$ \\[0.5em]
  $^{1}$i14 \qquad
  $^{2}$University of Melbourne \qquad
  $^{3}$University of California, Santa Cruz
}

\begin{document}

\maketitle

\begin{abstract}
Mean cross-entropy is the standard validation metric for language models, but it can fail to track model quality during training. We examine this in two common scenarios. First, in Qwen2.5-1.5B SFT on synthetic fact-learning, we find that mean CE rises substantially after the initial learning phase while held-out fact-recall accuracy remains near its peak. Second, we find that in top-\(K\) distillation on TinyStories, decreasing \(K\) improves median CE while worsening mean CE; the Top-5 student attains the highest LLM-judge score and crosses below its teacher on median CE, despite having the worst mean CE. In both cases, median CE correlates much more closely with task performance than does mean CE.

\smallskip
Analyzing how bulk and tail percentile CE move during training reveals that training reshapes the empirical per-token CE distribution. In top-\(K\) distillation, smaller \(K\) yields a distribution with more mass at both extremes, decreasing the median and increasing the mean. In Qwen SFT, the bulk saturates quickly while the tail extends in the latter half of training. In both, the task-evaluation metric appears more sensitive to the bulk than to the tail.

\smallskip
Practically, we recommend reporting a small set of percentile CE summaries alongside the mean, and using concordance among them as a tool to keep track of distribution reshaping, as well as a low-cost diagnostic for when mean and median CE disagree on model selection.
\end{abstract}
\section{Introduction}
\label{sec:intro}

Mean cross-entropy~\citep{bengio2003neural} (equivalently, perplexity) is the standard validation metric for language models. But it is well documented that mean CE does not fully explain variation in downstream task performance~\citep{fan2024longppl, ruan2024scaling}. Indeed, it can fail when the empirical per-token CE distribution reshapes---meaning that the bulk and tail move differently---in ways that mass-averaging hides. This paper is motivated by the following question: \emph{when does mean CE fail to predict downstream model quality, and how can we detect this cheaply?}

Let $\ell_t = -\log q_\theta(x_t \mid x_{<t})$ denote the per-token loss, where $q_\theta$ is the model's output distribution. When most parts of the empirical per-token CE distribution improve together during training, mean CE and percentile summaries tend to give the same model-selection\footnote{By \emph{model selection} we mean choosing among candidate checkpoints or training objectives to optimize performance on a specific task.} signal. But in practice, we see more nuanced behavior --- for example, the upper tail grows while typical-token prediction improves. To detect this, we can use \emph{percentile CE} summaries, which are robust statistics. We use the median as our \emph{bulk percentile}; it tracks \emph{typical-token loss} without being unduly affected by a small number of high-loss tokens. We use $p_{95}$ as our \emph{tail percentile}.

We study how mean and percentile CE summaries interact in two scenarios that arise frequently in practice: supervised fine-tuning and top-\(K\) distillation, in which the student is trained to match the teacher's top-$K$ token probabilities, renormalized to sum to one. We use a notion of concordance to test whether different CE summaries give consistent model-selection signals. This lets us distinguish regimes where mean CE is a reliable proxy for task performance from regimes where the choice of CE summary becomes consequential.

\paragraph{Notation.}
The empirical per-token CE distribution is the collection $\{\ell_t\}_{t=1}^T$ over evaluated validation tokens indexed by $t=1,\dots,T$. The \emph{mean CE} is $\bar{\ell} = \tfrac{1}{T}\sum_t \ell_t$. The \emph{$k$-th percentile CE}, denoted $p_k$, is the value below which $k\%$ of per-token CE losses fall; for example, $p_{50}$ is the median. We reserve $q$ (with subscripts as needed) for model output distributions.

\paragraph{Concordance regimes.}
Let $\mathcal{S}$ denote a set of CE summaries. Throughout most of the paper, we will take $\mathcal{S}=\{\mathrm{mean},\mathrm{median}, p_{95}\}$. Given a family $\mathcal M$ of model checkpoints, the \emph{concordance} of $\mathcal{S}$ over $\mathcal{M}$, denoted by $\pi(\mathcal{S})$, is the degree to which the summaries in $\mathcal{S}$ agree on how pairs of checkpoints are ranked relative to one another (see Appendix~\ref{app:dir-align} for the precise definition).\footnote{We will sometimes take $\mathcal{S}=\{\mathrm{mean},\mathrm{median}\}$ in order to compare mean and median CE directly. In that case, $\pi(\mathcal{S})$ is essentially the same as Kendall's notion of concordance (see Appendix~\ref{app:dir-align}).}

Intuitively, when concordance is high, the summaries in $\mathcal{S}$ tell the same selection story, in which case selecting models on the basis of mean CE alone is less likely to conflict with selecting by bulk or tail percentiles. When concordance is low, model selection is more likely to depend on whether the target evaluation is bulk- or tail-sensitive. In this way, we view concordance as a complement to mean and percentile CE that can help detect \emph{when} the choice of CE summary becomes consequential.

Motivated by the experiments that follow, we identify three regimes:
\begin{enumerate}
    \item \textbf{Directionally uniform improvement}: high $\pi(\mathcal{S})$, where CE summaries in $\mathcal{S}$ give consistent model-selection signals.
    \item \textbf{Deliberate reshaping}: low $\pi(\mathcal{S})$ induced by a training objective, e.g.\ top-$K$ distillation.
    \item \textbf{Implicit reshaping}: low $\pi(\mathcal{S})$ induced by training dynamics, e.g.\ during SFT.
\end{enumerate}
Our main contributions are as follows:
\begin{itemize}
    \item We find that Qwen2.5-1.5B SFT fact-learning (Section~\ref{sec:comprehension}) falls under regime 3: fine-tuning dynamics implicitly reshape the per-token CE distribution in a way not explained by mean CE alone. Recall accuracy is much more closely aligned with median CE than mean CE.

    \item We find that top-\(K\) self-distillation on TinyStories (Section~\ref{sec:distillation}) falls under regime 2: decreasing \(K\) corresponds to greater reshaping of the loss distribution, moving mass toward both extremes (Section~\ref{subsec:result2}) and decreasing median CE while increasing mean CE. Students can cross below their teacher on median CE (Section~\ref{subsec:result1} and Appendix~\ref{app:finewebedu}).

    \item We find that LLM-judge scores on the quality of subsequent generation are much more closely aligned with median CE than mean CE (Section~\ref{subsec:result1}), with the Top-5 student achieving the highest judge score despite having the worst mean CE. By contrast, when top-$K$ distilled students are replaced by controls of various sizes and learning rates, the cohort falls under regime 1 (Section~\ref{subsec:subsets}), where mean CE and median CE agree as model selectors.
\end{itemize}

The TinyStories findings above replicate qualitatively on linear attention  (Appendix~\ref{app:gla}).
\section{Related Work}
\label{sec:related}

\paragraph{Perplexity and its limitations.}
\citet{fan2024longppl} show that perplexity correlates poorly with long-context benchmark score since it averages uniformly over all tokens. They propose LongPPL, which selectively weights key tokens. We identify a complementary failure: uniform averaging lets a heavy tail of high-loss tokens dominate the mean. Where LongPPL learns a token-importance criterion, we summarize the full loss distribution via percentiles---a simpler, task-agnostic diagnostic. \citet{ruan2024scaling} show that pretraining-loss scaling is not uniformly predictive of downstream behavior. MAUVE~\citep{pillutla2021mauve} and LLM-as-judge~\citep{zheng2023judging} are alternative quality measures; we show that even within the CE framework, the choice of summary statistic matters.

\paragraph{Token typicality and generation.}
\citet{meister2023typical} propose local typical sampling: selecting tokens whose information content lies near the model's conditional entropy. Our percentile-CE perspective is conceptually related---both treat the ``typical'' region of the predictive distribution as governing quality. \citet{holtzman2020curious} motivate nucleus sampling on a similar basis.

\paragraph{Distillation and tail treatment.}
\citet{hinton2015distilling} introduced knowledge distillation, with LM-specific variants in DistilBERT~\citep{sanh2019distilbert}, TinyBERT~\citep{jiao2020tinybert}, and MiniLLM~\citep{gu2024minillm}. \citet{dasgupta2026tail} propose a tail-aware divergence that \emph{increases} the tail's contribution to KL, arguing standard KL underweights the tail for faithful distribution matching. We target a different criterion (bulk-aligned practical quality) and show that \emph{reducing} the tail's contribution via top-$K$ truncation can improve LLM-judged generation quality even while worsening mean CE.

\paragraph{Robust aggregation under heavy-tailed losses.}
Robust alternatives to the mean (e.g., median-of-means) achieve sub-Gaussian concentration under minimal moment assumptions~\citep{lugosi2019mean}. In training, \citet{zhang2018generalized} propose truncated CE losses for noisy labels---a conceptual parallel to our observation that high-loss tokens can dominate mean CE. Our percentile diagnostics instantiate this robust-aggregation view for language-model validation: median CE exposes typical-token behavior, while $p_{95}$ monitors the tail.

\section{A Fine-Tuning Case Study: Qwen Fact-Learning}
\label{sec:comprehension}

\subsection{Method} We generate 120 novel facts about fictional entities and express each as 30 Q/A phrasings, split into 20 train and 10 test per fact (2{,}400 train pairs and 1{,}200 held-out test pairs total). We fine-tune Qwen2.5-1.5B~\citep{qwen2025} with LoRA~\citep{hu2022lora} ($r{=}4$, $\alpha{=}8$) for 12 epochs. CE loss is computed on the gold answer string and is therefore decoding-independent; accuracy is measured both by greedy substring-match and by sampling pass@1 ($T{=}0.7$, top-$k{=}40$, $N{=}5$). (See Appendix~\ref{app:qwen-diagnostics} for more details.) 

\subsection{Results}
\begin{figure}[!h]
    \centering
    \includegraphics[width=\linewidth]{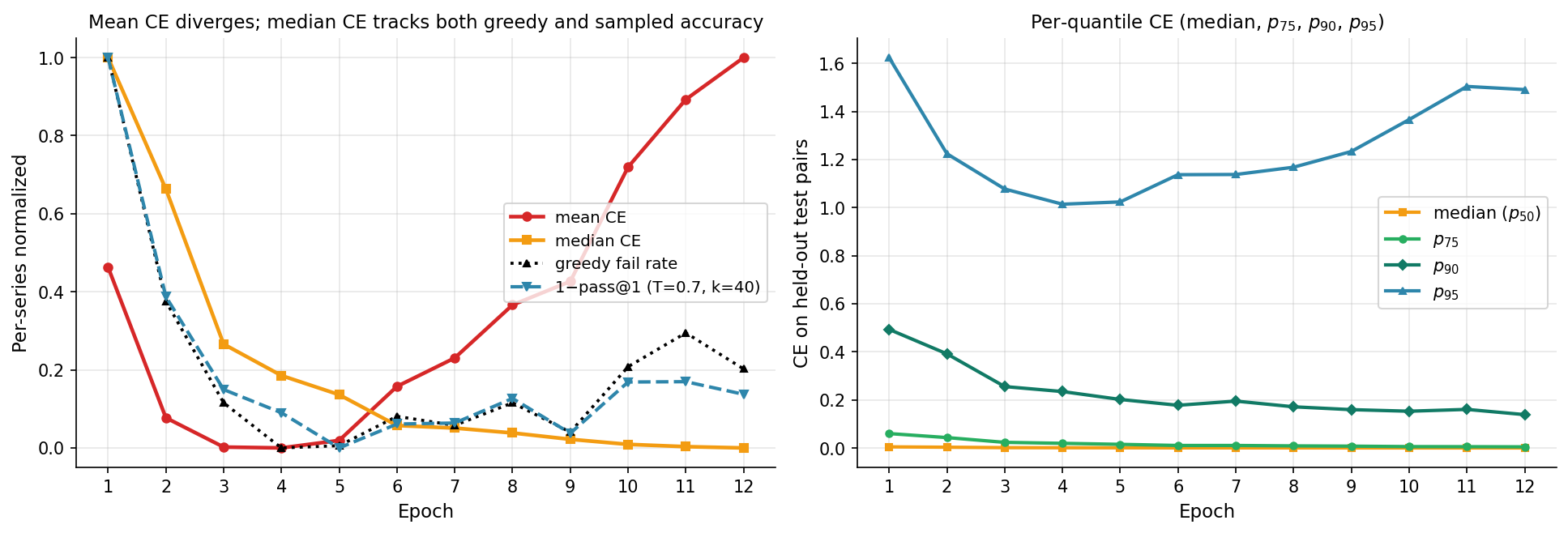}
    \caption{Qwen2.5-1.5B LoRA fine-tuning on 120 novel facts (1{,}200 held-out test pairs). \textbf{Left}: per-series normalized mean CE (red) and median CE (orange) alongside greedy substring-match fail-rate (black, dotted) and $1{-}$pass@1 (blue, dashed). After early improvement, mean CE rises while median CE falls to floor; both fail-rates track median CE far more closely than mean CE. \textbf{Right}: percentile CE summaries showing the bulk falls monotonically while $p_{95}$ traces a U-shape.}
    \label{fig:divergence}
\end{figure}

The observed trajectory exhibits qualitatively different behavior in early versus later epochs that mean CE alone obscures (Figure~\ref{fig:divergence}) -- an instance of \emph{implicit reshaping}, regime~3 from Section~\ref{sec:intro}.

\textbf{Median and mean diverge after early learning.} Through epoch 5, every CE summary moves consistently with both accuracy metrics. Between epochs 5 and 12, the picture shifts: median CE is at floor ($\approx 0.0001$) and $p_{95}$ rises $47\%$ ($1.01 \to 1.49$), while mean CE follows the tail, climbing $60\%$ to $0.346$. Greedy and pass@1 fail rates tick up only $\sim 3$pp each ($0.124 \to 0.152$ and $0.150 \to 0.176$). \emph{In either case, the mean CE rise is not accompanied by a comparable drop in task accuracy.}

\textbf{Median CE tracks the accuracy trajectory; mean CE does not.} Across the 12 training epochs, median CE has Pearson correlation $r{\geq}{+}0.84$ with both greedy and pass@1 fail rates (Table~\ref{tab:qwen-pearson}). Mean CE, by contrast, has weak correlations: $r{=}{+}0.27$ vs.\ greedy fail rate and $r{=}{+}0.12$ vs.\ pass@1 fail rate. \emph{Same trajectory under either decoding strategy; mean CE misses it under both.}

\begin{table}[H]
\centering
\small
\setlength{\tabcolsep}{6pt}
\caption{Pearson $r$ between CE summaries and fail-rates across the 12 Qwen training epochs (1{,}200 test pairs). Both fail-rate metrics (greedy substring-match and sampling pass@1 at $T{=}0.7$, top-$k{=}40$, $N{=}5$) are tracked far more strongly by median CE (and by $p_{90}$ and $p_{95}$) than by mean CE.}
\label{tab:qwen-pearson}
\begin{tabular}{lcc}
\toprule
& \multicolumn{2}{c}{\textit{across-epoch Pearson $r$ vs.\ fail rate}} \\
\cmidrule(lr){2-3}
CE summary & greedy & 1$-$pass@1 \\
\midrule
mean CE      & $+0.27$ & $+0.12$ \\
median CE    & $\mathbf{+0.84}$ & $\mathbf{+0.91}$ \\
$p_{90}$     & $+0.80$ & $+0.88$ \\
$p_{95}$     & $+0.76$ & $+0.65$ \\
$p_{99}$     & $-0.09$ & $-0.25$ \\
\bottomrule
\end{tabular}
\end{table}

\textbf{Tail interpretation.}
As seen in this experiment, the region of the loss distribution most aligned with task behavior can shift during training, in this case from the bulk to the upper tail. The mean-median divergence is visible in concordance: $\pi(\{\mathrm{mean}, \mathrm{median}\}) = 0.18$ across the 12 epochs, so mean and median CE disagree on which checkpoint is better in over 80\% of epoch pairs. Appendix~\ref{app:qwen-diagnostics} gives a more detailed discussion, including tail examples, correct/incorrect CE splits, and checkpoint-selection tables.

\section{Controlled Intervention: Top-$K$ Self-Distillation}
\label{sec:distillation}

The Qwen case study suggests that the informative region of the loss distribution can shift from the bulk in early training to the upper tail in later training. The following experiment, conducted on the TinyStories corpus, is structurally different: we use top-$K$ distillation to directly truncate the teacher tail signal, giving a controlled way to reshape the student's loss distribution while holding model size, data, and architecture fixed---isolating one controlled source of bulk--tail redistribution, namely the truncation level $K$. This is an instance of \emph{deliberate reshaping}, regime~2 from Section~\ref{sec:intro}.

\subsection{Method}

We work with the TinyStories~\citep{eldan2023tinystories} corpus, for two reasons. First, it lets us test whether the bulk-tail phenomenon appears at small scale: TinyStories models with \(\sim\)10--50M parameters can generate fluent stories, enabling controlled same-size teacher--student experiments and LLM-as-judge evaluation across many variants at modest compute. Second, its simplified vocabulary and grammar make the per-token CE distribution easier to interpret, while preserving the heavy-tailed structure of next-token prediction. \citet{eldan2023tinystories} show that sub-10M-parameter models can generate fluent text on TinyStories, a capability that otherwise requires much larger models on standard corpora. (See Appendix~\ref{app:finewebedu} for a replication at 150M on FineWebEdu.)

We train a baseline using token-based learning for 250K steps on a TinyStories corpus (8-layer transformer, 384-dim, 6 heads, ${\sim}$34M params; batch size $32$, sequence length $256$, giving $8{,}192$ tokens per step and $\sim 2.05$B tokens total). We freeze this baseline and use it as the teacher for all KL variants, distilling into students of \emph{identical architecture and size} for 250K steps. The only variable among students is the distillation objective. In top-$K$ KL distillation, the loss function is
\[
\mathcal{L}_{\mathrm{top\text{-}K}} = \KL(\tilde{q}_T^{(K)} \| q_S),
\]
where $q_T$ and $q_S$ are the teacher and student output distributions and $\tilde{q}_T^{(K)}$ retains only the top-$K$ teacher probabilities, renormalized to sum to one.
The student does not receive a KL gradient signal from teacher probability mass outside the retained top-$K$ set. 

Our primary quality indicator is an \textbf{LLM-as-judge}: 200 continuations scored by Claude Sonnet 4 (1--5 scale) on coherence, grammar, creativity, and overall quality, of which we report the overall score.

\subsection{Result 1: LLM-Judge scores track median CE; monotonic variation across $K$.}\label{subsec:result1} As $K$ decreases from full vocabulary to 5, the bulk improves (median CE drops from 0.632 to 0.525) while the tail worsens ($p_{95}$ CE rises from 5.73 to 7.82). Mean CE---a uniform average over the heavy-tailed distribution---tracks the worsening tail, rising from 1.489 to 1.708. We find that LLM-judge scores track the bulk: among distilled variants, Top-5 KL achieves the highest judge score (2.06) despite having the worst mean CE. While the absolute judge scores are modest, we use them for relative comparison within a matched model family.

\begin{figure}[!h]
    \centering
    \includegraphics[width=\linewidth]{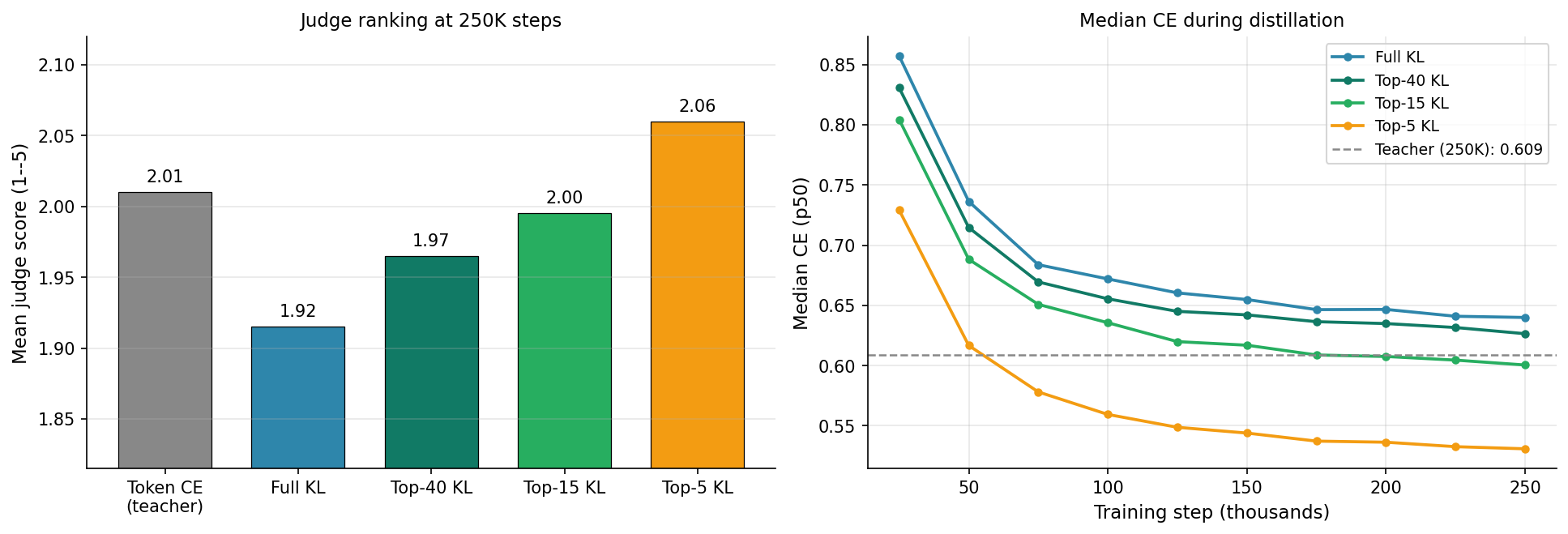}
    \caption{\textbf{Left}: mean judge scores for the five variants at 250K steps over the 200 evaluation prompts (significance claims rely on paired bootstraps in the main text). Top-5 KL has the highest mean judge score (2.06). \textbf{Right}: median CE during self-distillation, evaluated every 25K steps. Dashed line: the frozen teacher's median CE at its 250K endpoint (0.609). Top-5 KL crosses below the teacher's endpoint by $\sim$75K steps and continues to improve; Top-15 KL crosses near the end.}
    \label{fig:judge-and-crossing}
\end{figure}

Figure~\ref{fig:judge-and-crossing} visualizes the judge ranking (left) and the median-CE trajectory during training (right, with checkpoints every 25K steps). The right panel shows a surprising pattern: a student with the same architecture and parameter count can drop below its teacher's eventual median CE---Top-5 KL crosses below at $\sim$75K steps, while Top-15 KL crosses near the end of training ($\sim$175K). 

Note that this does \emph{not} mean the crossing students have surpassed the teacher as likelihood models (for example, they have worse mean CE; see Table~\ref{tab:tradeoff}), but rather that they achieve lower loss on the typical-token region of the validation distribution.

\begin{table}[!h]
\centering
\caption{TinyStories self-distillation at 250K steps. The frozen Token-CE baseline serves as teacher for same-size KL students (${\sim}$34M parameters). Decreasing $K$ improves median CE while worsening mean and $p_{95}$ CE, producing a controlled monotonic median--mean divergence. Judge scores are mean ratings over 200 prompts; significance claims for variant comparisons rely on paired-bootstrap differences, not on overlap of absolute-score intervals.}
\label{tab:tradeoff}
\small
\setlength{\tabcolsep}{5pt}
\begin{tabular}{llcccc}
\toprule
Model & Training objective & Mean CE & Median CE & $p_{95}$ CE & Judge \\
\midrule
\emph{Teacher} & \emph{Token CE (250K)} & \emph{1.442} & \emph{0.609} & \emph{5.57} & \emph{2.01} \\
\midrule
Student & Full KL & 1.489 & 0.632 & 5.73 & 1.92 \\
Student & Top-40 KL & 1.520 & 0.620 & 5.88 & 1.97 \\
Student & Top-15 KL & 1.568 & 0.596 & 6.37 & 2.00 \\
Student & Top-5 KL & \textbf{1.708} & \textbf{0.525} & \textbf{7.82} & \textbf{2.06} \\
\bottomrule
\end{tabular}
\end{table}

Unlike in the Qwen case study, where the divergence comes from training dynamics, top-$K$ distillation truncates the teacher tail signal directly while holding other variables fixed. The mean and median both change monotonically with $K$, corresponding to greater bulk-tail reshaping.

\subsection{Result 2: Top-$K$ Distillation Induces Loss Redistribution.}\label{subsec:result2} 
\begin{figure}[!h]
    \centering
    \includegraphics[width=\linewidth]{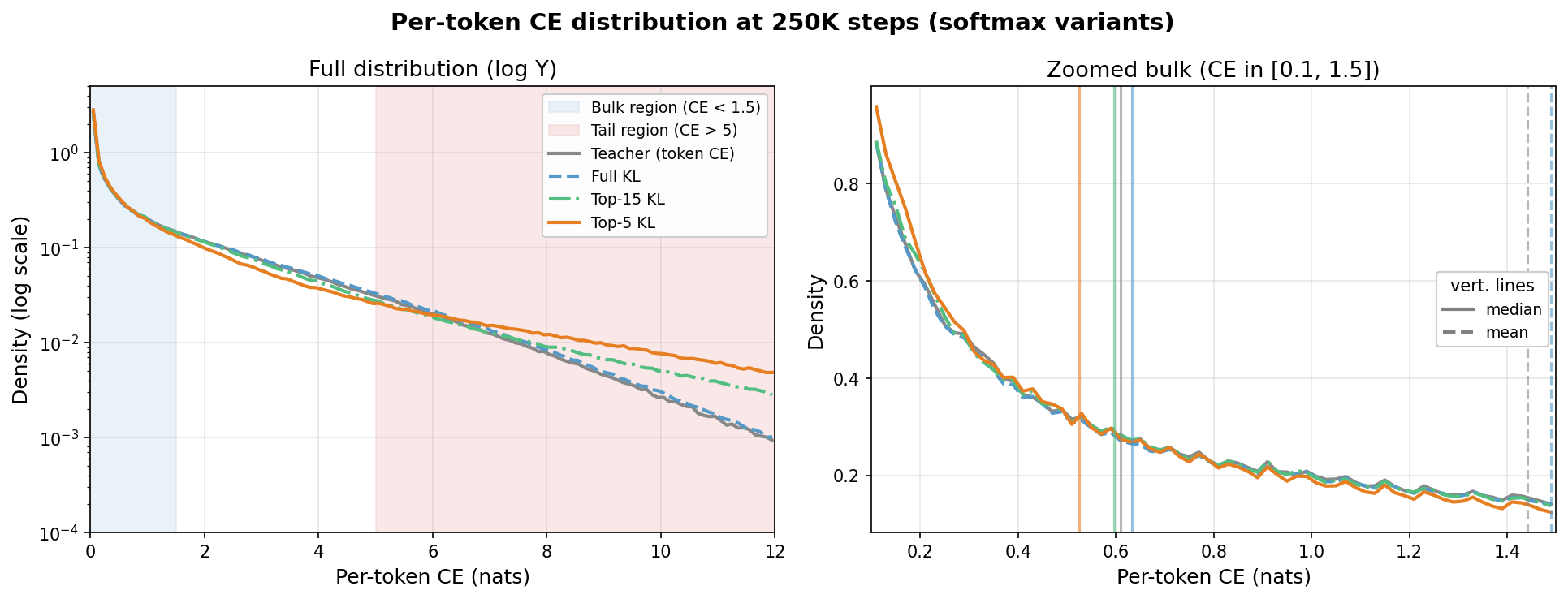}
    \caption{Per-token CE distributions at 250K steps for the four key variants. \textbf{Left} (log-scale Y, full range): the blue-shaded region (CE $<$ 1.5) is the high-density bulk where most tokens live; the red-shaded region (CE $>$ 5) is the rare high-loss tail that dominates mean CE. Top-5 KL's tail decays visibly more slowly than the others through the red-shaded region. \textbf{Right} (linear Y, zoomed bulk on CE $\in [0.1, 1.5]$): Top-5 KL has higher density near CE $\approx 0.1$--$0.3$ and lower density around CE $\approx 0.6$--$1.0$ than other methods. Solid vertical lines mark the median per method and dashed lines the mean. Top-5 KL has the lowest median ($0.525$) but the highest mean ($1.708$); the teacher has the highest median ($0.609$) but the lowest mean ($1.442$). 
    }
    \label{fig:distributions}
\end{figure}

Figure~\ref{fig:distributions} illustrates the distributional mechanism behind the above results: decreasing $K$ \emph{polarizes} the loss distribution. Comparing across $K$ at the end of training (250K steps), we see that smaller-$K$ students have more mass at both extreme CE ends and less in the middle band.\footnote{Appendix~\ref{app:redistribution} quantifies this and traces how it emerges \emph{during} training for the Top-5 student.}
 This is reflected by the median falling with the growing low-CE bulk, while the mean is \emph{dragged up} by the heavier tail.

\paragraph{Replication across different architectures.} The observations on the mean-median divergence and LLM-judge correlation with median are not specific to softmax attention: see Appendix~\ref{app:gla} for the results of an analogous experiment done with linear attention. 

\paragraph{Smaller students outperform teacher on typical-token prediction at larger scale.} We replicated at larger scale the effect of top-$K$ students crossing the teacher on median CE---see Appendix~\ref{app:finewebedu}. In that case, the students are 150M-parameter models and the teacher is a 355M GPT-2 medium model trained on FineWebEdu.

\subsection{Result 3: Mean CE Fails in the Low-Concordance Regime}
\label{subsec:subsets}

In order to test the hypothesis that mean CE fails to predict model quality when the downstream evaluator is sensitive to ``non-uniform" changes in token-loss distribution, we compare the family of distilled model checkpoints with a family of control checkpoints trained from scratch using different model sizes and learning rates. 

\begin{figure}[!h]
    \centering
    \includegraphics[width=\linewidth]{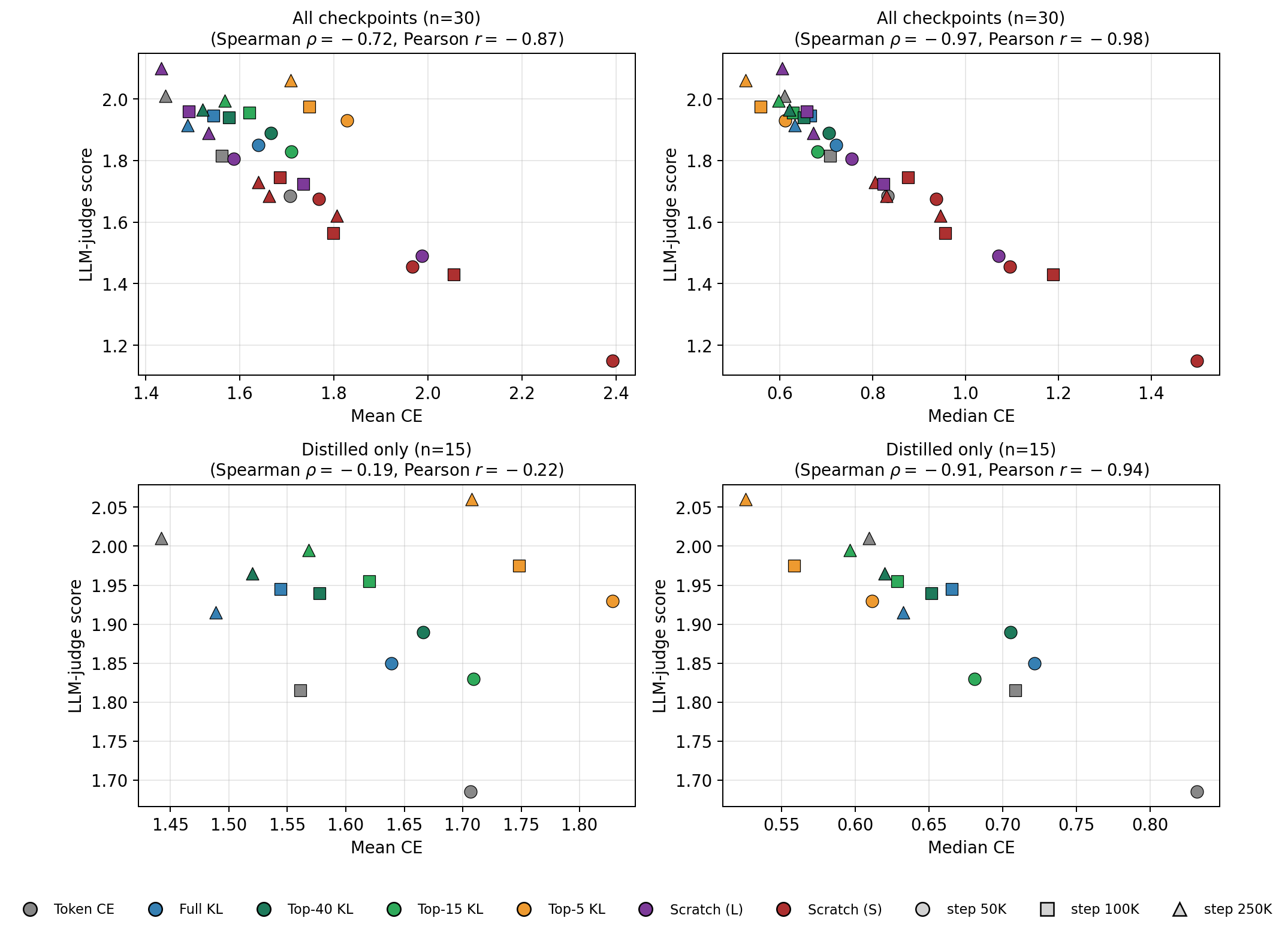}
    \caption{Per-checkpoint scatter of mean and median CE against LLM judge score. \textbf{Top row}: all 30 checkpoints. \textbf{Bottom row}: 15 distilled-only checkpoints. \textbf{Left column}: mean CE. \textbf{Right column}: median CE. Color indicates training family; marker shape indicates checkpoint step. The percentile advantage is most visible within the distilled subset (bottom row): mean CE shows essentially no monotonic trend with judge score, while median CE produces a clean descending curve.}
    \label{fig:scatter-ce-judge}
\end{figure}

In total, we compare 30 checkpoints from 10 independent training runs. All models are trained on TinyStories using the same training pipeline; the distilled family shares an architecture, while the from-scratch control family additionally varies model size and learning rate. Each run is saved as a separate checkpoint at 50K, 100K, and 250K steps. We separate them into two families:
\begin{itemize}
    \item \textbf{Distillation family} (5 runs, 15 checkpoints): Token CE (teacher), Full KL, Top-$K$ KL for $K \in \{5, 15, 40\}$.
    \item \textbf{From-scratch control family} (6 runs, 18 checkpoints): 2 model sizes $\times$ 3 learning rates, token-based learning, no distillation.
\end{itemize}
The Token CE run appears in both. See Appendix~\ref{app:inventory} for model sizes and learning rates. We find that (see Table~\ref{tab:subsets} and Figure~\ref{fig:scatter-ce-judge}):
\begin{itemize}
	\item \textbf{Within distilled checkpoints, mean CE collapses} ($\rho = -0.186$, $r = -0.217$)---essentially no rank power---while median CE holds ($\rho = -0.911$, $r = -0.935$); an instance of regime 2.
	\item \textbf{Within non-distilled controls, mean CE is essentially perfect} ($\rho = -0.977$, $r = -0.977$) and matches the best percentile ($\rho = -0.979$, $r = -0.982$); when models differ only in size, learning rate, or training duration, CE summaries broadly agree in direction and ranking; an instance of regime 1.
\end{itemize}
\begin{table}[!h]
\centering
\caption{Correlation between mean CE / median CE and LLM-judge score within subsets of the 30 softmax checkpoints ($\rho$ = Spearman, $r$ = Pearson). Mean CE matches median CE on the non-distilled controls but collapses on the distilled subset. See Table~\ref{tab:subsets-full} (Appendix~\ref{app:full-correlations}) for the full per-subset breakdown including the best-correlated percentile (softmax and linear-attention rows side by side).}
\label{tab:subsets}
\small
\setlength{\tabcolsep}{6pt}
\begin{tabular}{lccccc}
\toprule
Subset & $n$ & Mean $\rho$ & Mean $r$ & Median $\rho$ & Median $r$ \\
\midrule
All 30 & 30 & $-0.716$ & $-0.868$ & $-0.974$ & $-0.978$ \\
\midrule
Distilled only & 15 & $-0.186$ & $-0.217$ & $-0.911$ & $-0.935$ \\
Non-distilled (controls $+$ Token CE) & 18 & $-0.977$ & $-0.977$ & $-0.977$ & $-0.976$ \\
\bottomrule
\end{tabular}
\end{table}

\textbf{Concordance between mean and median.} Why does mean CE work for non-distilled controls and fail for distilled variants? One answer is given by concordance. Within the 18 non-distilled controls (the 6 control runs above, including the Token-CE teacher, $\times$ 3 step snapshots), mean and median CE agree on $96\%$ of checkpoint pairs; within the 15-checkpoint distillation family this drops to $51\%$, i.e.\ $\pi(\{\mathrm{mean}, \mathrm{median}\}) = 0.96$ vs.\ $0.51$. Taking the larger set of summaries $\mathcal{S}=\{\mathrm{mean}, \mathrm{median}, p_{95}\}$ shows the same pattern, with $\pi(\mathcal{S}) = 0.92$ vs.\ $0.41$ in the non-distilled vs.\ distilled families respectively. (For more details, see Appendix~\ref{app:dir-align}, Table~\ref{tab:pi-S}.) 

In the low-concordance regime, selection depends on whether the target evaluation is closer to the typical-token bulk or to the high-loss tail; as shown above, in this case LLM-judged quality follows the bulk percentiles rather than mean CE.

Appendix~\ref{app:shape} explains the distributional source of this concordance collapse: top-$K$ variants show upper-tail expansion after robust standardization, while non-distilled and Token-CE-distilled checkpoints have similar standardized percentile profiles. (pairwise $\ell_2$ $\sim 0.25$ within scratch, $0.27$ between scratch and Token-CE-distilled).

\begin{figure}[!h]
    \centering
    \includegraphics[width=0.9\linewidth]{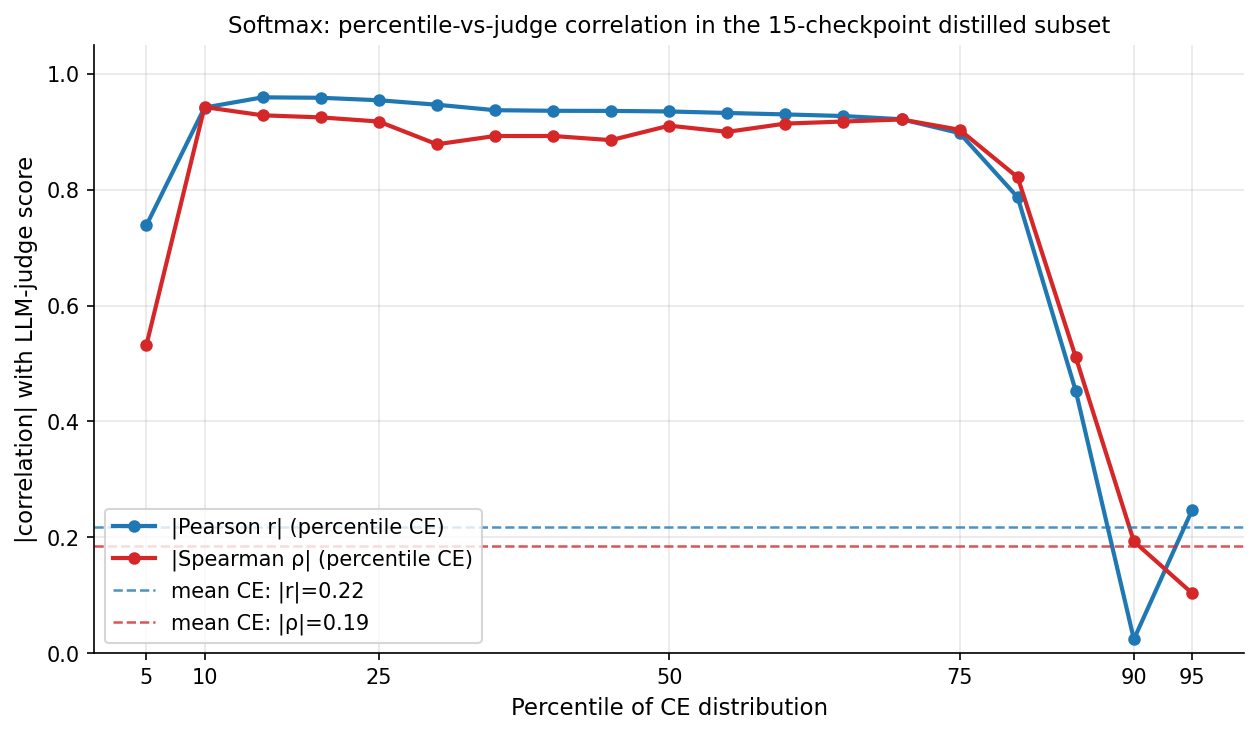}
    \caption{Correlation between CE percentile and LLM judge score within the 15-checkpoint distilled subset (Token-CE teacher + four KL students across 3 training steps). Blue: $|$Pearson $r|$. Red: $|$Spearman $\rho|$. Dashed: mean-CE correlation, $|r| = 0.217$, $|\rho| = 0.186$. Median CE reaches $|r| = 0.935$, $|\rho| = 0.911$; the tail ($p_{85}$--$p_{95}$) is unreliable. More details in Appendix~\ref{app:full-correlations}.}
    \label{fig:percentile-curve}
\end{figure}

\textbf{Other percentile summaries.} Across all 30 checkpoints, median CE reaches Pearson $|r| = 0.978$ and Spearman $|\rho| = 0.974$, while mean CE achieves $|r| = 0.868$ and $|\rho| = 0.716$ (full per-percentile table for both architectures in Appendix~\ref{app:full-correlations}). Within the 15-checkpoint distilled subset, the gap widens: mean CE collapses to $|r| = 0.217$, $|\rho| = 0.186$, while median CE holds at $|r| = 0.935$, $|\rho| = 0.911$, and the best bulk percentile reaches $|r| = 0.960$ ($p_{15}$), $|\rho| = 0.943$ ($p_{10}$) (Figure~\ref{fig:percentile-curve}). The relevant failure is not a single magic percentile but summary-statistic disagreement under reshaping: bulk percentiles agree with judge quality while mean CE follows the tail. 

Thus the result is not that a single percentile is uniquely optimal. Rather, we see from Figure \ref{fig:percentile-curve} that almost any bulk percentile from $p_{10}$ to $p_{80}$ ranks models much better than does mean CE, while extreme high percentiles ($p_{90}$, $p_{95}$) deteriorate as they become tail-dominated. This supports using a small percentile summary as a diagnostic rather than tuning a particular percentile.

\section{Discussion: When to Use Mean CE}
\label{sec:discussion}

\paragraph{Model selection.}
Median CE can be a better model selector than mean CE, depending on the downstream evaluation objective. In the distillation experiment, the low-concordance distillation family is the regime where the choice of CE summary becomes consequential (see Table~\ref{tab:model-selection}), in contrast with the high-concordance control family where mean CE suffices. The Top-5 vs.\ Full KL judge difference is statistically significant (paired bootstrap on the 200 prompts: $+0.145$, 95\% CI $[+0.06, +0.24]$); the Top-5 vs.\ teacher difference is within bootstrap uncertainty.

\begin{table}[!h]
\centering
\caption{Model selection within the 15-checkpoint distillation family. Mean CE selects the Token-CE teacher, while median CE and the judge select Top-5 KL.}
\label{tab:model-selection}
\small
\setlength{\tabcolsep}{5pt}
\begin{tabular}{lccccc}
\toprule
Selection rule & Selected checkpoint & Mean CE & Median CE & $p_{95}$ CE & Judge \\
\midrule
Best mean CE & Token CE 250K (teacher) & 1.442 & 0.609 & 5.57 & 2.01 \\
Best median CE & Top-5 250K & 1.708 & 0.525 & 7.82 & \textbf{2.06} \\
Best judge & Top-5 250K & 1.708 & 0.525 & 7.82 & \textbf{2.06} \\
\bottomrule
\end{tabular}
\end{table}

\paragraph{Training-dynamics interpretation.}
Median and other percentile CEs can help even in interpreting a single training trajectory. In the Qwen fine-tuning case study, mean CE indicates classical overtraining during later epochs, with validation mean CE rising substantially ($\sim 60\%$), while accuracy remains high ($\sim 3$pp drop). In this case, percentile CEs give a more precise account: the median has saturated while the upper tail continues to accumulate loss, and the rise in mean CE is not matched by a comparable degradation in task accuracy. Thus, if the goal is to understand training dynamics, mean CE alone can mislead. 

\paragraph{Practical recommendation.}
Report mean CE together with median and \(p_{95}\), and check concordance across checkpoints before running an external judge or downstream benchmark. The two percentile summaries have distinct roles: the median tracks the center of the typical-token bulk, while \(p_{95}\) monitors the high-loss tail. High concordance indicates that validation summaries are likely to agree; low concordance calls for a more deliberate choice of summary statistic for model selection and for more nuanced training-trajectory interpretation. In our distillation experiments, judged quality follows the bulk; rare-token fidelity or calibration may require tail-sensitive selection.

Note that percentile CE is not a universal replacement for mean CE. Mean CE remains the appropriate metric when the target is likelihood, distributional fidelity, or rare-token calibration, and it works well empirically when all summaries tell essentially the same story. However, reporting percentile CEs and concordance is a low-cost way to keep track of how the loss distribution changes during training, particularly for heavy-tailed empirical loss distributions. As shown in our experiments, divergence between mean and median CE is a real effect that can inform both training diagnostics and model selection.

\paragraph{Limitations.}
The above experiments use TinyStories or synthetic facts at scales up to 1.5B parameters; claims concern \emph{relative} model selection, not frontier-scale generation. The 30-checkpoint validation uses softmax attention (Appendix~\ref{app:gla} adds a linear-attention robustness check).
\section{Conclusion}
Even in common language-model training scenarios such as SFT and knowledge distillation, mean CE (equivalently, perplexity) alone can be an incomplete descriptor of model quality. Its reliability as a proxy for task performance depends on how the empirical loss distribution evolves and on its concordance with percentile CE summaries such as the median. We therefore recommend reporting a small percentile sweep and concordance alongside mean CE---both to track how the loss distribution reshapes during training and as a low-cost diagnostic for when model selection may depend on the choice of CE summary.
\vfill
\pagebreak
\bibliography{references}
\bibliographystyle{plainnat}

\FloatBarrier
\appendix

\setlength{\textfloatsep}{3pt plus 1pt minus 1pt}
\setlength{\intextsep}{2pt plus 1pt minus 1pt}
\setlength{\floatsep}{3pt plus 1pt minus 1pt}
\setlength{\abovecaptionskip}{3pt}
\setlength{\belowcaptionskip}{8pt}
\captionsetup[table]{skip=8pt}
\setlength{\parskip}{1pt}
\makeatletter
\renewcommand\paragraph{\@startsection{paragraph}{4}{\z@}{1.5ex \@plus 0.3ex \@minus 0.2ex}{-0.5em}{\normalfont\normalsize\bfseries\color{black}}}
\makeatother

\newpage
\section{Concordance}
\label{app:dir-align}

We give a more precise discussion of concordance. Let
\(\mathcal{M}\) be a finite family of model checkpoints, and let \(\mathcal{S}\)
be a set of \(\mathbb{R}\)-valued CE summary statistics of the empirical
per-token loss distribution. Given two distinct checkpoints
\(A,B\in\mathcal{M}\), we say that \(\mathcal{S}\) is
\emph{concordant on} \(\{A,B\}\) if
\[
\operatorname{sign}\,\!\bigl(s(A)-s(B)\bigr)
=
\operatorname{sign}\,\!\bigl(s'(A)-s'(B)\bigr)
\qquad
\text{for all } s,s'\in\mathcal{S},
\]
i.e. all summaries in \(\mathcal{S}\) rank \(A\) and \(B\) in the same way.

We define the \emph{concordance of \(\mathcal{S}\) over \(\mathcal{M}\)},
denoted by \(\pi(\mathcal{S})\in[0,1]\), to be the fraction of unordered
checkpoint pairs \(\{A,B\}\subset\mathcal{M}\) on which \(\mathcal{S}\) is
concordant. When \(\mathcal{S}=\{a, b\}\) contains exactly two summaries, \(\pi(\mathcal{S})\) is the standard Kendall concordance fraction $(1+\tau_{a,b})/2$, where $\tau_{a,b}$ is Kendall's \(\tau\) between the rankings induced by \(a\) and \(b\) over \(\mathcal{M}\).

\paragraph{Empirical \(\pi(\mathcal{S})\) on TinyStories checkpoints.}
Set \(\mathcal{S}=\{\mathrm{mean},\mathrm{median},p_{95}\}\). In
Section~\ref{subsec:subsets}, concordance values split sharply by checkpoint
family, as shown in the following table.

\begin{table}[H]
\centering
\small
\setlength{\tabcolsep}{3pt}
\caption{First column: concordance \(\pi(\mathcal{S})\) across checkpoint families. Other columns: pairwise concordance \(\pi(\mathcal{T})\), where \(\mathcal{T}\) ranges over two-element subsets of \(\mathcal{S}\).}
\label{tab:pi-S}
\begin{tabular}{lcccc}
\toprule
Checkpoint family \(\mathcal{M}\) & \(\pi(\mathcal{S})\) & \(\{\mathrm{mean},\mathrm{median}\}\) & \(\{\mathrm{mean},p_{95}\}\) & \(\{\mathrm{median},p_{95}\}\) \\
\midrule
All 30                      & 0.62 & 0.77 & 0.85 & 0.62 \\
Non-distilled (controls $+$ Token CE) (18) & 0.92 & 0.96 & 0.96 & 0.92 \\
Teacher + KL students (15)  & 0.41 & 0.51 & 0.90 & 0.41 \\
KL students only (12)       & 0.36 & 0.46 & 0.91 & 0.36 \\
\bottomrule
\end{tabular}
\end{table}

There is a sharp contrast between the non-distilled controls,
where \(\pi(\mathcal{S})=0.92=141/153\) pairs of checkpoints, and the distillation family, where
\(\pi(\mathcal{S})=0.41=43/105\) pairs. Restricting further to the KL students gives
\(\pi(\mathcal{S})=0.36\). In the terminology of Section~\ref{sec:intro}, this
matches the distinction between directionally uniform improvement and deliberate
reshaping.
\smallskip

The remaining columns in Table~\ref{tab:pi-S} localize this disagreement. Within
the distillation family, the two lowest pairwise concordance values are
\[
\pi(\{\mathrm{median},p_{95}\})=0.41,
\qquad
\pi(\{\mathrm{mean},\mathrm{median}\})=0.51.
\]
Thus, in nearly half of distillation-family checkpoint pairs, mean CE and median
CE disagree on which checkpoint is better. This gives a numerical expression of
the bulk-tail reshaping discussed in the main text.

\vfill
\newpage
\section{Qwen Diagnostics}
\label{app:qwen-diagnostics}

This appendix expands on the results of the Qwen fact-learning case study from Section~\ref{sec:comprehension}.

\paragraph{Correct vs.\ incorrect CE split.}
For each epoch we compute mean CE separately on examples answered correctly vs.\ incorrectly under greedy decoding (Table~\ref{tab:correct-incorrect}). After the early-training improvement (epoch~1$\to$4--5), mean CE rises on \emph{both} subsets, while the ratio  $\mathrm{CE}(\text{incorrect})/\mathrm{CE}(\text{correct})$ stays within $\sim 2.5$--$2.7$. Incorrectly decoded examples therefore consistently carry $\sim 2.6\times$ the target-token loss of correct ones, even as accuracy stays in a narrow $\sim 4$pp band.

\begin{table}[H]
\centering
\caption{Held-out test pairs split by greedy correctness. Mean CE rises on both correct and incorrect examples, while CE(incorrect)/CE(correct) stays near $\sim 2.6$.}
\label{tab:correct-incorrect}
\begin{tabular}{rcccc}
\toprule
Epoch & Accuracy & Mean CE (correct) & Mean CE (incorrect) & Ratio (incorrect/correct) \\
\midrule
1 & 0.732 & 0.243 & 0.605 & 2.49 \\
4 & 0.881 & 0.202 & 0.541 & 2.68 \\
5 & 0.891 & 0.208 & 0.559 & 2.68 \\
7 & 0.877 & 0.238 & 0.648 & 2.73 \\
10 & 0.859 & 0.296 & 0.742 & 2.51 \\
12 & 0.861 & 0.322 & 0.823 & 2.55 \\
\bottomrule
\end{tabular}
\end{table}

\paragraph{Tail examples.}
At epoch 12, the 20 highest-loss test examples are dominated by completion-style phrasings where the model generates ``\texttt{?}''. Examples include \textit{``Complete: Above Nimbusglade floats \_\_\_''} (expected: Skyhold) and \textit{``By official designation, Zephyria's capital is \_\_\_''} (expected: Crystalport). At the blank position, the model assigns low probability to the expected answer token and prefers punctuation such as ``\texttt{?}'', so CE evaluated against the gold answer string spikes sharply, even though the same facts are answered correctly under other phrasings. The failure is therefore format-specific rather than knowledge loss.

\paragraph{Mean CE and accuracy are miscalibrated.}
Mean CE is a reasonable \emph{directional} indicator of fail rate in this case study: both bottom around epochs 3--5 and rise afterward. What it misses is the \emph{magnitude} of the late-training divergence. From the minimum at epoch 4 to epoch 12, mean CE rises $60\%$ from \(0.216\) to \(0.346\), while accuracy stays within a narrow \(\sim 4\)pp band of its peak. Percentiles explain the mismatch: median CE is at floor by epoch~6 ($\le 4{\times}10^{-4}$), so typical-token behavior is essentially unchanged, while the upper tail expands ($p_{95}$ rises $47\%$, $p_{99}$ rises $77\%$). Thus the mean-CE increase reflects reshaping---bulk saturation plus tail growth---rather than a proportionate degradation in task performance. Within the early plateau, mean CE is also essentially flat across epochs 3--5 (range $< 0.003$), whereas $p_{95}$ differentiates the plateau and selects epoch~4, matching the best-accuracy checkpoint (Table~\ref{tab:qwen-selection}).

\begin{table}[H]
\centering
\caption{Checkpoint selection under different criteria (v8c data, $1{,}200$ test pairs). Mean CE, $p_{95}$ CE, and accuracy all select epoch 4; median CE selects epoch 12 and is the only rule that diverges from accuracy, due to post-saturation drift (median keeps decreasing toward floor while accuracy drifts down).}
\label{tab:qwen-selection}
\begin{tabular}{lccccc}
\toprule
Selection rule & Selected epoch & Mean CE & Median CE & $p_{95}$ CE & Accuracy \\
\midrule
Best mean CE  & 4  & \textbf{0.216} & 0.00094 & 1.014 & 0.877 \\
Best median CE & 12 & 0.346 & \textbf{0.00010} & 1.491 & 0.848 \\
Best $p_{95}$ CE   & 4 & 0.216 & 0.00094 & \textbf{1.014} & 0.877 \\
Best accuracy & 4 & 0.216 & 0.00094 & 1.014 & \textbf{0.877} \\
\bottomrule
\end{tabular}
\end{table}

\begin{figure}[H]
    \centering
    \includegraphics[width=\linewidth]{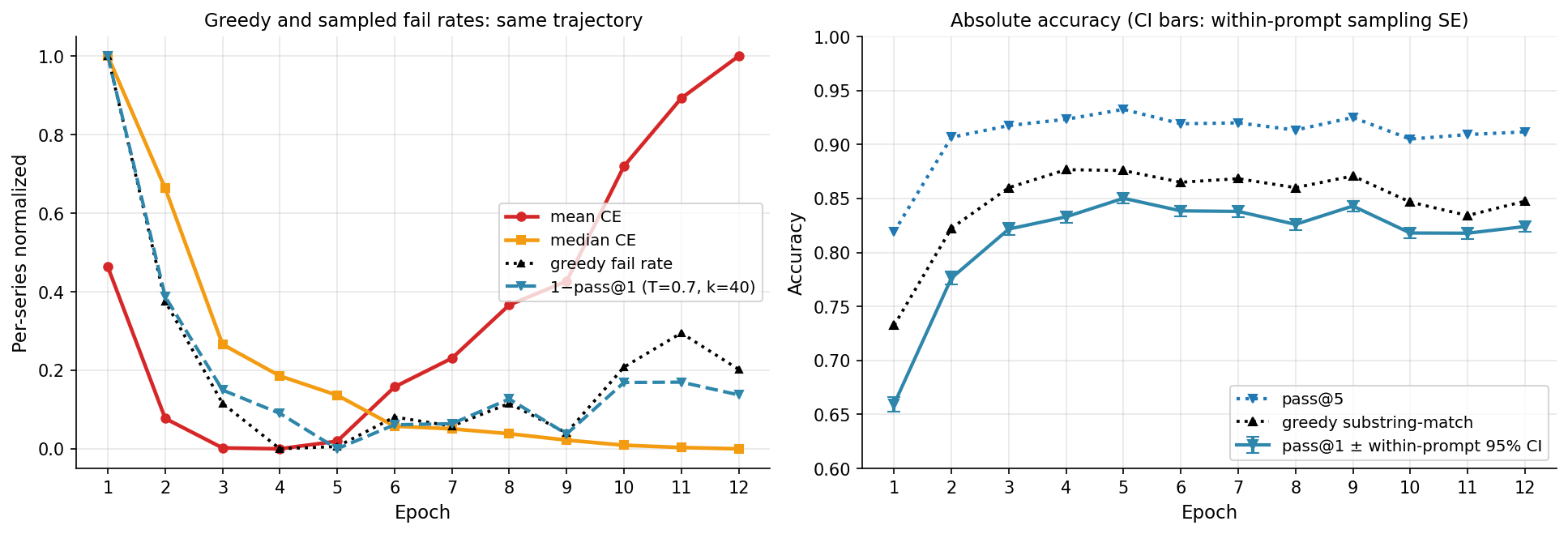}
    \caption{\textbf{Left}: (same as Figure~\ref{fig:divergence}, left) per-series normalized fail-rate trajectories for greedy substring-match (black, dotted) and pass@1 sampling (blue, dashed), alongside normalized mean and median CE. \textbf{Right}: absolute accuracy values. Greedy substring-match and pass@1 both peak around epoch~4--5 in the $0.83$--$0.88$ range and drift to $0.82$--$0.85$ by epoch~12. Pass@5 (any-of-5 sampled correctness) sits in $0.82$--$0.93$ and shows the same shape. Error bars on pass@1 are within-prompt 95\% CIs from the $N{=}5$ sampling variance ($\leq{\pm}0.7\%$).}
    \label{fig:qwen-v8c-absolute}
\end{figure}

\paragraph{Greedy and sampled accuracy, same trajectory.} A natural worry about Section~\ref{sec:comprehension}'s greedy substring-match accuracy is that the bulk-tail divergence is an artifact of greedy decoding. The CE statistics that drive our analysis are computed from $q_\theta(x_t \mid x_{<t})$ on the gold answer string and are therefore decoding-independent; only the accuracy measurement could in principle change. To check, we evaluate Monte Carlo sampling-based accuracy at every epoch on the same 1{,}200-pair test set: at each adapter checkpoint we generate $N{=}5$ samples per prompt with $T{=}0.7$, top-$k{=}40$, and count substring matches---reporting both pass@1 (mean fraction-correct over samples) and pass@5 (any-of-$N$ correct). Figure~\ref{fig:divergence}~(left) already overlays the greedy and pass@1 fail-rates with the mean and median CE curves; Figure~\ref{fig:qwen-v8c-absolute} shows absolute accuracy values, with within-prompt 95\% confidence intervals on pass@1 derived from the per-pair sampling variance ($\hat{p}_i(1{-}\hat{p}_i)/N$ aggregated over 1{,}200 prompts; CI half-width $\leq{\pm}0.7\%$ at every epoch). Greedy and sampled accuracy show the same qualitative trajectory: rapid early improvement through epoch~4--5, then slow decline of $\sim 3$pp by epoch~12, while mean CE rises $60\%$ over the same interval. The bulk-tail divergence is therefore not a greedy-decoding artifact and is robust under sampling.

\vfill
\newpage
\section{TinyStories Protocol and Checkpoints}
\label{app:inventory}

\subsection*{LLM-as-judge protocol}\label{app:judge-protocol}

\paragraph{Prompts and generation.}
We draw 200 prompts once from the TinyStories validation set (seed 42), using 64-token prefixes. This prompt set is used for every checkpoint. Each checkpoint generates 128 new tokens per prompt with temperature \(0.8\) and top-\(k=40\) sampling, using fixed seeds for reproducibility. All checkpoints use the same hyperparameters.

\paragraph{Judge.}
Continuations are scored by Claude Sonnet 4 on four 1--5 absolute-scale criteria: \emph{coherence}, \emph{grammar}, \emph{creativity}, and \emph{overall}. Throughout the paper, we report the \emph{overall} score, averaged across prompts. The judge sees only the prompt prefix and continuation; no model identifier or training family is revealed. Each continuation is scored independently. Judging is absolute, not pairwise.

\paragraph{Reporting and judge biases.}
We report mean judge scores aggregated over the 200 prompts, without bootstrap confidence intervals on absolute per-checkpoint means. Because every variant is judged on the same prompt set, intervals on absolute scores would be dominated by prompt-difficulty heterogeneity (different prompts elicit different-quality responses across the board) rather than by uncertainty about each variant's quality, and would not function as paired-comparison significance markers; we therefore rely on the qualitative ranking and on monotonic patterns across $K$. LLM-as-judge scores can carry systematic biases such as length, repetition, or fluency-vs.-correctness preferences; we use a fixed judge model and prompt template across all checkpoints, so any systematic bias is shared.

\paragraph{Independent sanity check.}
As an independent sanity check, we also compute CBT (Children's Book Test \citep{hill2016goldilocks}) word-prediction accuracy across the 30 checkpoints; CBT correlates strongly with the judge ($r=+0.84$, $\rho=+0.71$), confirming that the judge ranking is not idiosyncratic.

\subsection*{Checkpoint inventory}

The 30 softmax-attention checkpoints used in Section~\ref{subsec:subsets} come from 10 independent training runs on TinyStories, each evaluated at 50K, 100K, and 250K steps, detailed in the following table.

\begin{table}[H]
\centering
\caption{All 10 training runs and checkpoints (distilled family: 1--5; from-scratch controls: 1, 6--10). All models here use softmax attention (8-layer, 384-dim, 6 heads, ${\sim}$34M params for ``Large''; 4-layer, 256-dim, 4 heads, ${\sim}$8M params for ``Small''). Runs 2--5 use Run 1 at 250K steps as teacher.}

\label{tab:inventory}
\begin{tabular}{cllccc}
\toprule
Run & Objective & $K$ / Notes & Size & Checkpoints (K steps) & $n$ \\
\midrule
1 & Token CE & lr=3e-4 (also serves as teacher) & Large & 50, 100, 250 & 3 \\
2 & Full KL & $K{=}V$ & Large & 50, 100, 250 & 3 \\
3 & Top-$K$ KL & $K{=}40$ & Large & 50, 100, 250 & 3 \\
4 & Top-$K$ KL & $K{=}15$ & Large & 50, 100, 250 & 3 \\
5 & Top-$K$ KL & $K{=}5$ & Large & 50, 100, 250 & 3 \\
\midrule
6 & Token CE & lr=1e-3 & Large & 50, 100, 250 & 3 \\
7 & Token CE & lr=1e-4 & Large & 50, 100, 250 & 3 \\
8 & Token CE & lr=1e-3 & Small & 50, 100, 250 & 3 \\
9 & Token CE & lr=1e-4 & Small & 50, 100, 250 & 3 \\
10 & Token CE & lr=3e-4 & Small & 50, 100, 250 & 3 \\
\midrule
\multicolumn{5}{r}{Total checkpoints:} & 30 \\
\bottomrule
\end{tabular}
\end{table}

\vfill
\newpage
\section{FineWebEdu: Same Bulk-Tail Reshaping at Larger Scale}
\label{app:finewebedu}

\begin{figure}[H]
    \centering
    \includegraphics[width=\linewidth]{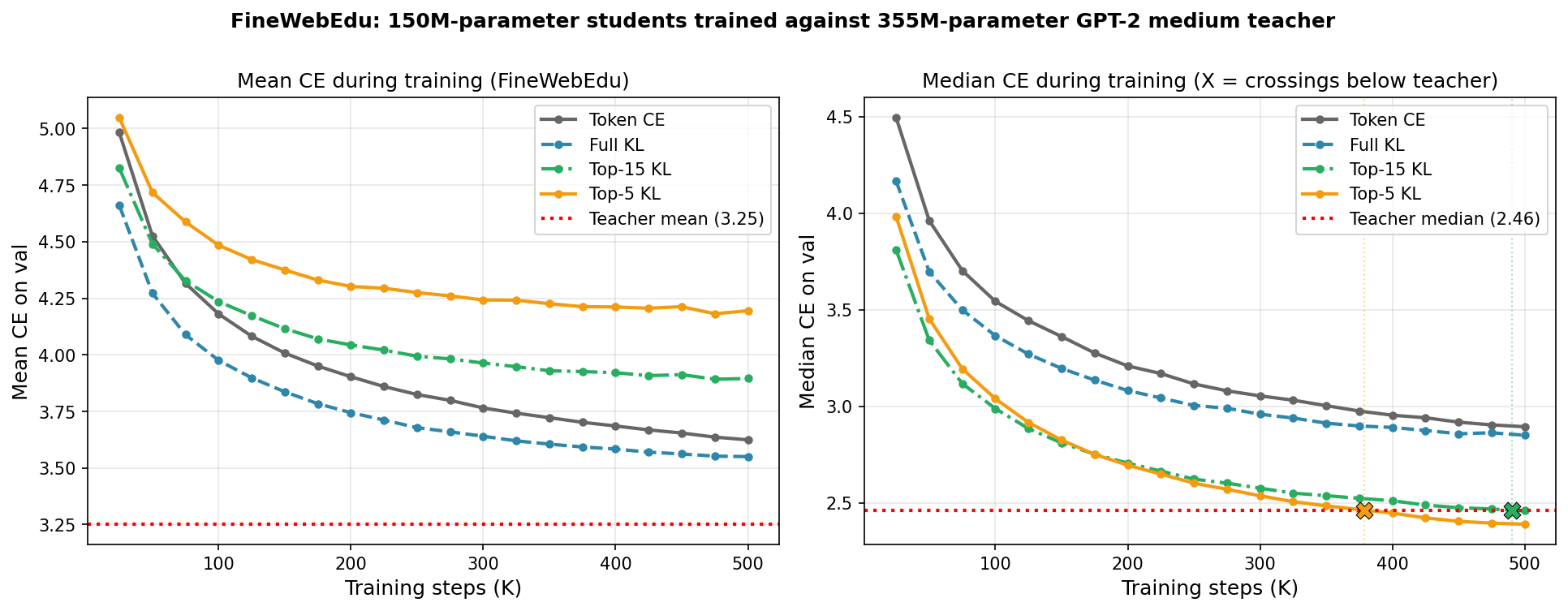}
    \caption{FineWebEdu: 150M-parameter students trained for 500K steps against a 355M-parameter GPT-2 medium teacher. \textbf{Left}: validation mean CE. All students remain above the teacher's mean ($3.25$, red dotted), and Top-5 KL is the worst on this metric. \textbf{Right}: validation median CE. Top-5 KL crosses below the teacher's median ($2.46$) at $\sim$379K steps; Top-15 KL reaches the teacher's median at $\sim$490K. Token CE and Full KL never cross.}
    \label{fig:finewebedu-curves}
\end{figure}

To test whether the TinyStories mean-vs.-median CE pattern persists at larger scale, we ran a 150M-parameter distillation experiment on FineWebEdu~\citep{penedo2024fineweb}, using a 355M-parameter GPT-2 medium teacher and training students for 500K steps (batch size $32$, sequence length $256$, giving $8{,}192$ tokens per step and $\sim 4.1$B tokens total). Figure~\ref{fig:finewebedu-curves} shows the validation mean and median CE trajectories for four students.
\smallskip

Notably, the Top-5 KL student crosses below the teacher's median CE at $\sim$379K steps and ends at 2.39, beating the teacher's median by 0.07 nats while being $2.4{\times}$ smaller (Table~\ref{tab:finewebedu}). 

\begin{table}[H]
\centering
\caption{FineWebEdu: 150M students cross below the 355M GPT-2 medium teacher on median CE, but not mean CE.}
\label{tab:finewebedu}
\begin{tabular}{lccc}
\toprule
Model & Params & Mean CE & Median CE \\
\midrule
\emph{Teacher (GPT-2 medium)} & \emph{355M} & \emph{3.25} & \emph{2.46} \\
\midrule
Token CE & 150M & 3.62 & 2.89 \\
Full KL & 150M & 3.55 & 2.85 \\
Top-15 KL & 150M & 3.90 & \textbf{2.46} \\
Top-5 KL & 150M & 4.20 & \textbf{2.39} \\
\bottomrule
\end{tabular}
\end{table}

\vfill
\newpage
\section{Mass Redistribution Under Top-$K$ KL}
\label{app:redistribution}

Section~\ref{sec:distillation} argues that top-$K$ distillation polarizes the per-token CE distribution. Table~\ref{tab:redistribution} quantifies this redistribution \emph{across $K$} at the 250K-step endpoint, relative to the teacher. As $K$ decreases, the moderate-loss band (CE $1.5$--$5$) drains; mass shifts toward both the very-low-CE bulk ($<0.5$) and the extreme tail ($>10$), with the effect strongest for Top-5 KL.

\begin{table}[H]
\centering
\caption{Mass redistribution from teacher to top-$K$ KL students at 250K steps. Entries are percentages of per-token CE samples in each region; $\Delta$ is in percentage points.}
\label{tab:redistribution}
\begin{tabular}{lcccccc}
\toprule
Method & $<0.1$ & $0.1$--$0.5$ & $0.5$--$1.5$ & $1.5$--$5$ & $5$--$10$ & $>10$ \\
\midrule
Teacher & 26.3\% & 20.4\% & 21.0\% & 25.8\% & 6.1\% & 0.4\% \\
Full KL & 25.9\% & 20.2\% & 20.6\% & 26.2\% & 6.5\% & 0.6\% \\
Top-15 KL & 26.4\% & 20.7\% & 20.8\% & 24.3\% & 6.2\% & 1.6\% \\
Top-5 KL & \textbf{27.8\%} & \textbf{21.4\%} & \textbf{20.0\%} & \textbf{20.7\%} & \textbf{7.2\%} & \textbf{2.9\%} \\
\midrule
$\Delta$ (Top-5 $-$ Teacher) & $+1.5$ & $+1.0$ & $-1.0$ & $\mathbf{-5.1}$ & $+1.1$ & $\mathbf{+2.5}$ \\
\bottomrule
\end{tabular}
\end{table}

\begin{figure}[H]
    \centering
    \includegraphics[width=0.88\linewidth]{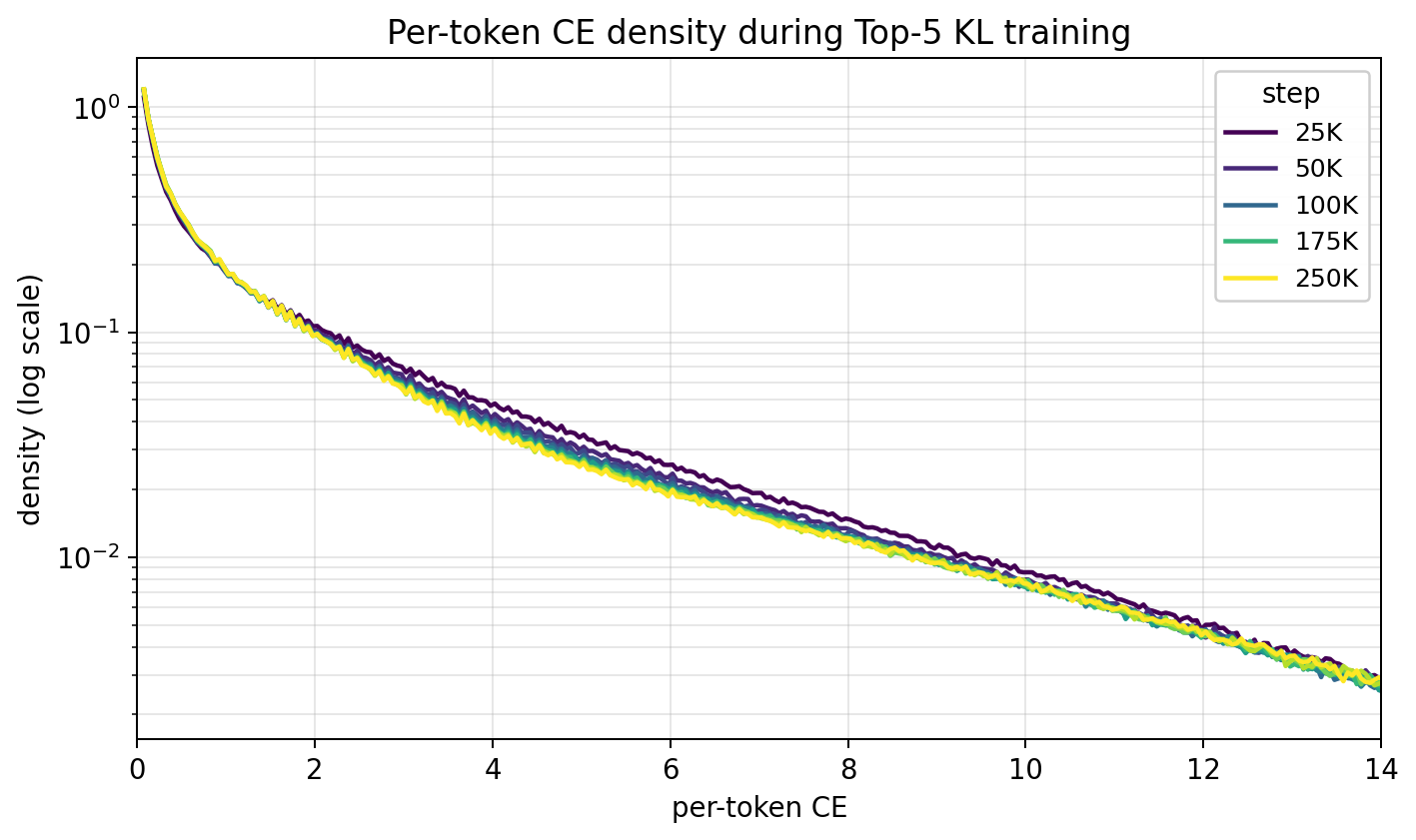}
    \caption{Per-token CE density evolution \emph{during} Top-5 KL training, evaluated on the full $6.2$M-token TinyStories validation set at every 25K-step checkpoint (log $y$, full range CE $\in[0, 14]$). The moderate-loss region (CE $\sim 1$--$8$) thins as training proceeds. The upper tail (CE $\gtrsim 10$) moves much less, drifting slightly upward in the last $\sim 75$K steps after the bulk has largely saturated.}
    \label{fig:top5-within-training-density}
\end{figure}

\paragraph{Within-training trajectory for Top-5.}
The redistribution above is across-condition (top-$K$ endpoints at 250K steps). To see whether the same mass-toward-extremes story holds \emph{within} a single distillation run, we examine the Top-5 KL student at every 25K-step checkpoint on the full $6.2$M-token TinyStories validation set (Figure~\ref{fig:top5-within-training-density}). From step 25K to 250K, bulk percentiles drop substantially ($p_{10}$: $0.0091\to 0.0057$, $-38\%$; median: $0.73\to 0.53$, $-27\%$), and the mass fraction at CE $\le 0.5$ grows from $43.8\%$ to $49.0\%$. Tail percentiles barely move ($p_{90}$: $-12\%$, $p_{95}$: $-4\%$), and $p_{99}$ inches up only marginally ($13.44\to 13.71$, $+2\%$), with most of the drift concentrated in the last $\sim 75$K steps---once the bulk has largely saturated, the extreme tail begins to creep outward. This echoes the late-training tail extension seen in the Qwen experiment. On the whole, the redistribution is dominated by bulk concentration, not bidirectional mass movement to both extremes.

\smallskip

This refines the across-$K$ picture: the heavier endpoint tail at smaller $K$ (Table~\ref{tab:redistribution}) reflects mostly \emph{failure to compress the tail} during training rather than active tail extension. With less teacher signal for low-probability tokens, the Top-5 student's tail compresses far less than its bulk does---and far less than the tail of higher-$K$ students---but the within-run motion is still in the compressing direction, not the extending direction.

\vfill
\newpage
\section{Distribution-Shape Diagnostics}
\label{app:shape}

The analysis in Section~\ref{subsec:subsets} tells us \emph{whether} CE summaries agree across the checkpoints; the diagnostics here help visualize what distributional reshaping \emph{looks like} when concordance is low.
\smallskip

Let $p_k^{(m)}$ denote the $k$-th percentile of the empirical per-token loss distribution $\{\ell_t\}$ from Section~\ref{sec:intro}, for a given model $m$. Denote the interquartile range by 
\[\mathrm{IQR}_m = p_{75}^{(m)} - p_{25}^{(m)}.\]
The \emph{standardized} $k$-th percentile,
\[
\widetilde{p}_k^{(m)}
= \frac{p_k^{(m)} - p_{50}^{(m)}}{\mathrm{IQR}_m},
\]
is the signed displacement of the $k$-th percentile from the median, measured in units of IQR. This puts models on a common robust scale, so that for a given $k$, differences in standardized $k$-th percentile across $m$ highlight changes in distributional shape. We consider two statistics:
\begin{itemize}
    \item \textbf{Standardized upper-tail percentile}: the scalar $\widetilde{p}_{95}^{(m)}$. A rise indicates that the upper tail extends farther relative to the bulk.
    \item \textbf{Standardized percentile profile}: the collection of $\widetilde{p}_k^{(m)}$ for $k \in \{5, 10, \dots, 95\}$.
    \end{itemize}
\begin{figure}[H]
    \centering
    \includegraphics[width=\linewidth]{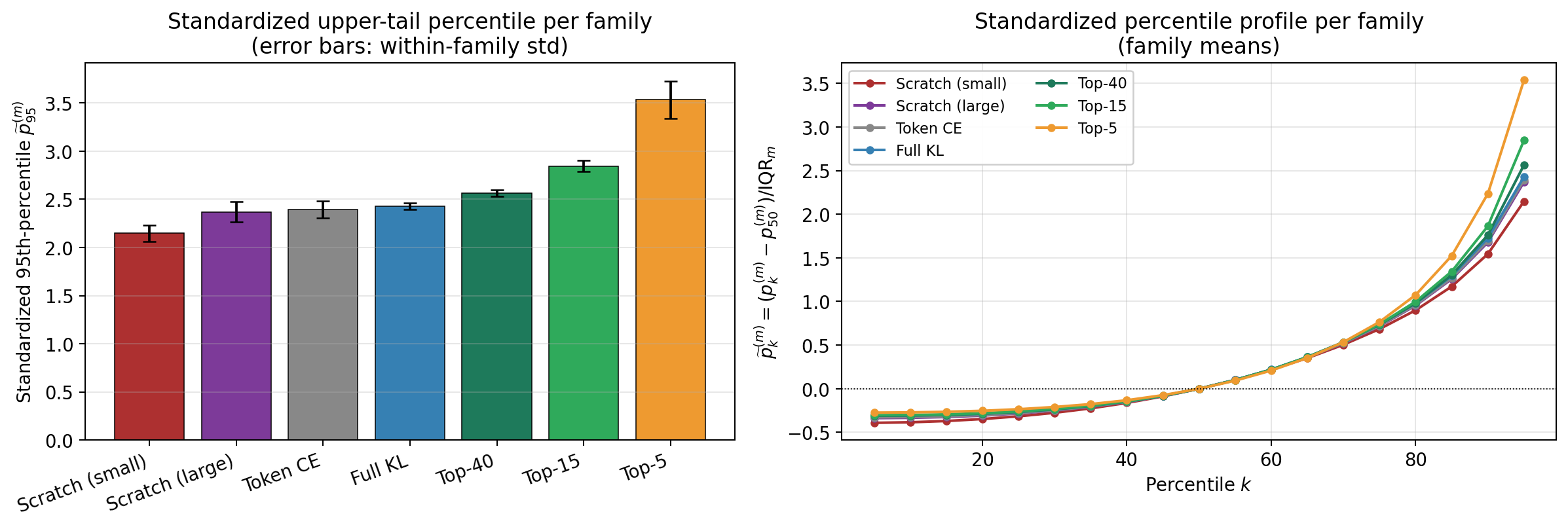}
    \caption{\textbf{Left}: standardized upper-tail percentile $\widetilde{p}_{95}^{(m)}$ per model family across all 30 checkpoints (error bars: within-family std).
    \textbf{Right}: standardized percentile profiles $\widetilde{p}_k^{(m)}$ averaged within family.
    }
    \label{fig:shape}
\end{figure}

They tell the same qualitative story (Figure~\ref{fig:shape}): the standardized upper-tail percentile $\widetilde{p}_{95}^{(m)}$ is $2.24 \pm 0.14$ across the scratch baselines and $2.39 \pm 0.09$ across the 3 Token-CE-distilled checkpoints. It rises monotonically with decreasing $K$ to $3.54 \pm 0.19$ for Top-5, a $\sim 58\%$ increase over the from-scratch baseline. 
\smallskip

To quantify shape differences between two models $m_1$ and $m_2$, we treat the percentile profile as a vector and compute the $\ell_2$ distance $d_2\bigl(\widetilde{p}^{(m_1)}, \widetilde{p}^{(m_2)}\bigr)$. We find that the mean pairwise $\ell_2$ distance is $0.25$ within the control family (over all 105 scratch--scratch pairs), $0.27$ between scratch and Token-CE-distilled checkpoints (45 pairs), and $1.51$ between scratch and Top-5 (45 pairs). Non-distilled and Token-CE checkpoints show little shape change after robust standardization, while top-$K$ variants increasingly diverge in the upper percentiles as $K$ decreases. This supports the interpretation that the rank disagreement between mean CE and bulk percentiles in the distilled subset is accompanied by bulk-tail redistribution, while in the non-distilled subset it is not.

\vfill
\newpage
\section{Robustness Check: a Linear-Attention Architecture}
\label{app:gla}

To test whether the bulk-tail reshaping observed on softmax attention generalizes, we repeat the controlled-intervention experiment (Section~\ref{sec:distillation}) and the percentile-correlation analysis (Section~\ref{subsec:subsets}) on a linear-attention architecture. Each transformer block replaces softmax self-attention with RetNet-style retention~\citep{sun2023retentive}: identity feature map with fixed per-head multi-scale decay $\gamma_h = 1 - 2^{-5-h}$, and a per-head LayerNorm before the output projection. For a given head $h$, the state recurrence and output are
\[
S_t^{(h)} = \gamma_h \, S_{t-1}^{(h)} + k_t^{(h)\top} v_t^{(h)}, \qquad o_t^{(h)} = q_t^{(h)}\, S_t^{(h)},
\]
where $S_t^{(h)} \in \mathbb{R}^{d_h \times d_h}$ is the per-head hidden state and $q_t^{(h)}, k_t^{(h)}, v_t^{(h)} \in \mathbb{R}^{d_h}$. Equivalently, the parallel form $o_i^{(h)} = \sum_{j \le i} \gamma_h^{\,i-j}\, (q_i^{(h)\top} k_j^{(h)})\, v_j^{(h)}$ applies a causal mask with geometric decay $\gamma_h^{\,i-j}$. Per-head outputs are passed through LayerNorm, concatenated, and projected. Model dimensions, training budget, data, optimizer, batch size, evaluation pipeline, and judge model all match the softmax setting (8-layer, 384-dim, 6 heads, ${\sim}$34M params, 250K steps on TinyStories).

\begin{figure}[H]
    \centering
    \includegraphics[width=0.64\linewidth]{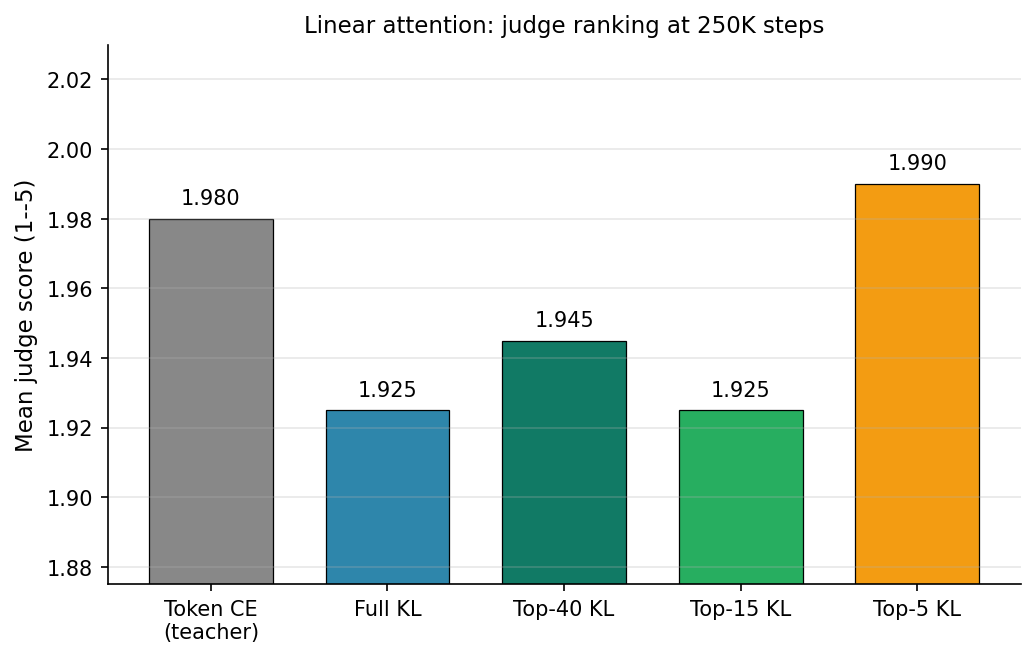}
    \caption{Linear attention: mean judge scores at 250K over 200 prompts. The same qualitative pattern as the softmax experiment (Figure~\ref{fig:judge-and-crossing}, left): Top-5 KL has the highest mean judge score among distilled variants, but the gaps among the KL variants are smaller than in the softmax architecture.}
    \label{fig:linear-judge}
\end{figure}

\begin{figure}[H]
    \centering
    \includegraphics[width=0.9\linewidth]{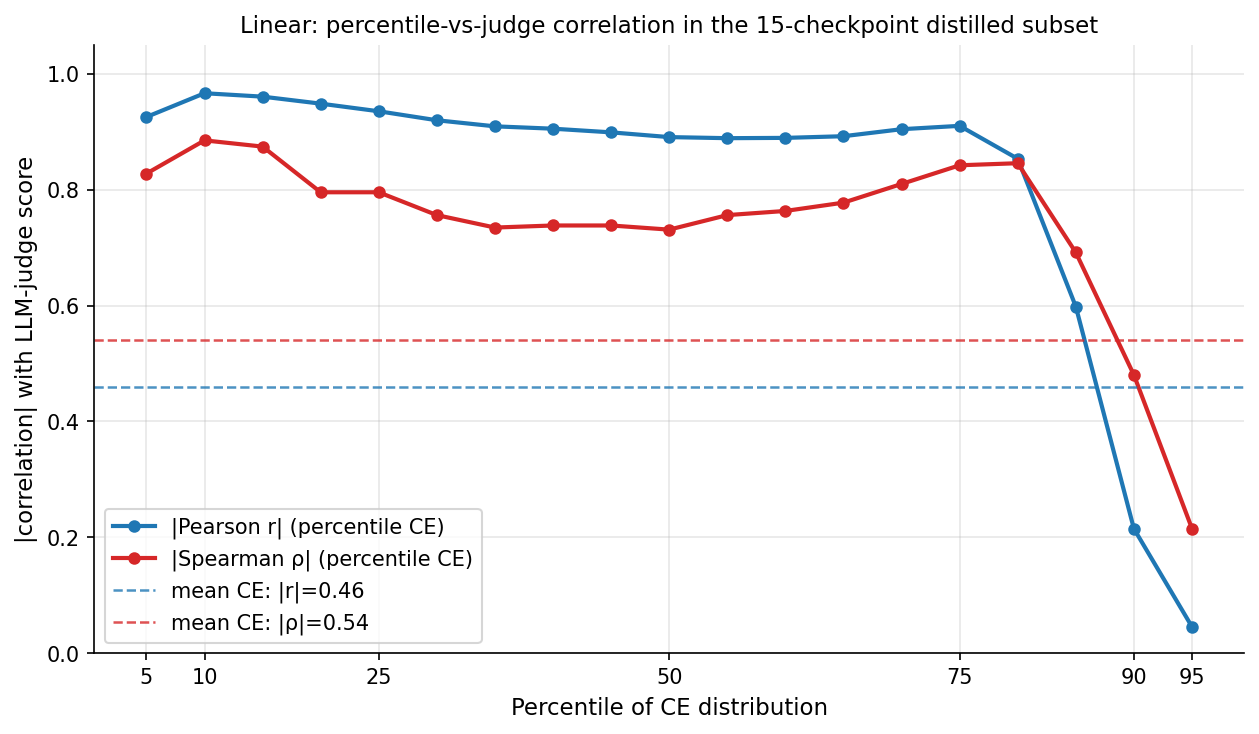}
    \caption{Linear attention: percentile-vs-judge correlation within the 15-checkpoint distilled subset (analogue of Figure~\ref{fig:percentile-curve}). Median CE reaches $|r| \approx 0.89$ while mean CE (dashed) sits well below at $|r| = 0.46$, $|\rho| = 0.54$. The same qualitative pattern as the softmax architecture: a wide gap between bulk-percentile correlations and mean-CE correlation in the distilled regime, with a tail breakdown beyond $p_{85}$.}
    \label{fig:linear-percentile-sweep}
\end{figure}

\paragraph{Monotonic dose-response at 250K (parallel to Table~\ref{tab:tradeoff}).}
Table~\ref{tab:gla-tradeoff} shows the same monotonic dose-response as softmax: as $K$ decreases, median CE drops while mean and $p_{95}$ CE rise, and Top-5 KL has the worst mean CE and the highest mean judge score among distilled variants. The relative changes closely match softmax (median: $-13.5\%$ vs.\ $-13.8\%$; mean: $+18.1\%$ vs.\ $+18.4\%$). The judge-score gaps are smaller, however (Top-5 vs.\ Full KL: $+0.065$ vs.\ $+0.145$ on softmax), so we report the qualitative rank replication rather than significance claims (Figure~\ref{fig:linear-judge}).
\begin{table}[H]
\centering
\small
\setlength{\tabcolsep}{5pt}
\caption{Linear-attention self-distillation on TinyStories at 250K steps, with the same architecture and parameter count for the teacher and students. Judge scores are mean ratings over the 200 evaluation prompts; we do not report unpaired bootstrap intervals on absolute means (see Appendix~\ref{app:judge-protocol}).}
\label{tab:gla-tradeoff}
\begin{tabular}{llcccc}
\toprule
Model & Training objective & Mean CE & Median CE & $p_{95}$ CE & Judge \\
\midrule
\emph{Frozen teacher} & \emph{Token CE (250K)} & \emph{1.488} & \emph{0.665} & \emph{5.64} & \emph{1.980} \\
\midrule
Student & Full KL & 1.529 & 0.699 & 5.74 & 1.925 \\
Student & Top-40 KL & 1.561 & 0.683 & 5.91 & 1.945 \\
Student & Top-15 KL & 1.610 & 0.652 & 6.42 & 1.925 \\
Student & Top-5 KL & \textbf{1.757} & \textbf{0.575} & \textbf{7.92} & \textbf{1.990} \\
\bottomrule
\end{tabular}
\end{table}
\paragraph{Subset correlations (parallel to Section~\ref{subsec:subsets}).}
We evaluate 30 unique linear-attention checkpoints drawn from the same training grid as the softmax cohort (Section~\ref{subsec:subsets}, Appendix~\ref{app:inventory}): 15 distilled (Token-CE teacher $+$ four KL students $\times$ 3 training steps) and 18 non-distilled (the 6-run control family, including the Token-CE teacher, $\times$ 3 steps; the Token-CE teacher is shared between the two subsets, giving 30 unique checkpoints). The six control runs match the softmax controls in model size and learning rate (2 sizes $\times$ 3 learning rates). The same regime separation as in softmax appears: on the distilled-only subset, mean CE collapses (Spearman $\rho = -0.541$, Pearson $r = -0.459$) while median CE holds substantially better ($\rho = -0.731$, $r = -0.891$); on the non-distilled controls both summaries are essentially indistinguishable ($\rho = -0.977$ for both, with $r = -0.985$ for mean and $r = -0.976$ for median). The mean-CE collapse is less severe than in softmax ($\rho = -0.541$ vs.\ $-0.186$), reflecting the narrower judge-score spread on this architecture (1.92--1.99 vs.\ 1.92--2.06). Figure~\ref{fig:linear-percentile-sweep} visualizes the bulk-vs-mean correlation gap; Table~\ref{tab:subsets-full} (Appendix~\ref{app:full-correlations}) reports the full per-subset breakdown including the best-correlated single percentile.

\paragraph{Comparison to softmax.}
The qualitative replication of the monotonic pattern in $K$, the model-selection tension within the distilled subset, and the percentile-vs-mean correlation gap together suggest that the bulk-tail reshaping observed in the main text is not specific to softmax attention. The judge-score gaps on linear attention are smaller and should be interpreted cautiously; broader architectural coverage is left to future work.

\vfill
\newpage
\section{Full Percentile Correlation Table (Both Architectures)}
\label{app:full-correlations}

This appendix tabulates the per-percentile Pearson and Spearman correlations between CE summaries and LLM-judge score that support the regime-separation claims in Section~\ref{subsec:subsets} (softmax) and Appendix~\ref{app:gla} (linear attention). 


\paragraph{Subset-level summary with best percentile.}
Table~\ref{tab:subsets-full} expands the compact mean-vs-median table in the main text (Table~\ref{tab:subsets}) with the strongest single percentile per subset, and reports softmax and linear-attention rows side by side.

\begin{table}[H]
\centering
\caption{Subset correlations between CE summaries and LLM-judge score, for both architectures ($\rho$ = Spearman, $r$ = Pearson; ``Best pct.'' columns report the best-correlated single percentile per subset; bolded rows are the regime-separation findings). Mean CE works well only on the non-distilled controls; in every other subset, bulk percentiles carry the rank signal.}
\label{tab:subsets-full}
\small
\setlength{\tabcolsep}{4pt}
\begin{tabular}{llccccc}
\toprule
Architecture & Subset & $n$ & Mean $\rho$ & Mean $r$ & Best pct.\ $\rho$ & Best pct.\ $r$ \\
\midrule
\multirow{3}{*}{Softmax}
  & All checkpoints & 30 & $-0.716$ & $-0.868$ & $-0.974$ (median) & $-0.982$ ($p_{65}$) \\
  \cmidrule(lr){2-7}
  & Distilled only & 15 & $\mathbf{-0.186}$ & $\mathbf{-0.217}$ & $\mathbf{-0.943}$ ($p_{10}$) & $\mathbf{-0.960}$ ($p_{15}$) \\
  & Controls & 18 & $\mathbf{-0.977}$ & $\mathbf{-0.977}$ & $\mathbf{-0.979}$ ($p_{40}$) & $\mathbf{-0.982}$ ($p_{75}$) \\
\midrule
\multirow{3}{*}{Linear}
  & All checkpoints & 30 & $-0.764$ & $-0.884$ & $-0.970$ ($p_{15}$) & $-0.982$ ($p_{70}$) \\
  \cmidrule(lr){2-7}
  & Distilled only & 15 & $\mathbf{-0.541}$ & $\mathbf{-0.459}$ & $\mathbf{-0.885}$ ($p_{10}$) & $\mathbf{-0.967}$ ($p_{10}$) \\
  & Controls & 18 & $\mathbf{-0.977}$ & $\mathbf{-0.985}$ & $\mathbf{-0.977}$ ($p_{80}$) & $\mathbf{-0.988}$ ($p_{80}$) \\
\bottomrule
\end{tabular}
\end{table}

\paragraph{Per-percentile correlations.}
Table~\ref{tab:full-percentile} reports Pearson $r$ and Spearman $\rho$ correlations between CE at every 5th percentile and LLM-judge score, computed (i) across all 30 checkpoints in each architecture and (ii) within the 15-checkpoint distilled subset (Token-CE teacher $+$ four KL students $\times$ 3 training steps). Within the distilled subset---where median-mean divergence is active---the gap between bulk percentiles and mean CE widens dramatically: mean-CE $|r|$ drops from $0.87$ to $0.22$ on softmax and from $0.89$ to $0.46$ on linear, while bulk percentiles continue to predict judged quality strongly. The extreme percentiles $p_{90}$ and $p_{95}$ become tail-dominated and unreliable on both architectures and in both subsets.

\begin{table}[H]
\centering
\footnotesize
\setlength{\tabcolsep}{3.5pt}
\caption{Pearson $r$ and Spearman $\rho$ between CE percentile and LLM-judge score, for both architectures, computed across all checkpoints and within the 15-checkpoint distilled subset. Bolded entries mark the strongest correlation per column.}
\label{tab:full-percentile}
\begin{tabular}{lcccc@{\hspace{1.5em}}cccc}
\toprule
 & \multicolumn{4}{c}{Softmax} & \multicolumn{4}{c}{Linear attention} \\
\cmidrule(lr){2-5} \cmidrule(lr){6-9}
 & \multicolumn{2}{c}{All (30)} & \multicolumn{2}{c}{Distilled (15)} & \multicolumn{2}{c}{All (30)} & \multicolumn{2}{c}{Distilled (15)} \\
\cmidrule(lr){2-3} \cmidrule(lr){4-5} \cmidrule(lr){6-7} \cmidrule(lr){8-9}
CE summary & $r$ & $\rho$ & $r$ & $\rho$ & $r$ & $\rho$ & $r$ & $\rho$ \\
\midrule
Mean   & $-0.868$ & $-0.716$ & $-0.217$ & $-0.186$ & $-0.884$ & $-0.764$ & $-0.459$ & $-0.541$ \\
$p_5$     & $-0.892$ & $-0.768$ & $-0.739$ & $-0.532$ & $-0.907$ & $-0.903$ & $-0.926$ & $-0.828$ \\
$p_{10}$    & $-0.936$ & $-0.954$ & $-0.942$ & $\mathbf{-0.943}$ & $-0.939$ & $-0.952$ & $\mathbf{-0.967}$ & $\mathbf{-0.885}$ \\
$p_{15}$    & $-0.947$ & $-0.961$ & $\mathbf{-0.960}$ & $-0.929$ & $-0.950$ & $\mathbf{-0.970}$ & $-0.961$ & $-0.875$ \\
$p_{20}$    & $-0.957$ & $-0.967$ & $-0.959$ & $-0.925$ & $-0.958$ & $-0.952$ & $-0.949$ & $-0.796$ \\
$p_{25}$    & $-0.962$ & $\mathbf{-0.974}$ & $-0.955$ & $-0.918$ & $-0.962$ & $-0.950$ & $-0.936$ & $-0.796$ \\
$p_{30}$    & $-0.965$ & $-0.969$ & $-0.947$ & $-0.879$ & $-0.965$ & $-0.951$ & $-0.920$ & $-0.756$ \\
$p_{35}$    & $-0.969$ & $-0.965$ & $-0.937$ & $-0.893$ & $-0.969$ & $-0.949$ & $-0.910$ & $-0.735$ \\
$p_{40}$    & $-0.973$ & $-0.966$ & $-0.936$ & $-0.893$ & $-0.973$ & $-0.948$ & $-0.905$ & $-0.738$ \\
$p_{45}$    & $-0.976$ & $-0.971$ & $-0.936$ & $-0.886$ & $-0.976$ & $-0.946$ & $-0.899$ & $-0.738$ \\
Median & $-0.978$ & $-0.974$ & $-0.935$ & $-0.911$ & $-0.978$ & $-0.946$ & $-0.891$ & $-0.731$ \\
$p_{55}$    & $-0.980$ & $-0.972$ & $-0.933$ & $-0.900$ & $-0.979$ & $-0.950$ & $-0.889$ & $-0.756$ \\
$p_{60}$    & $-0.982$ & $-0.974$ & $-0.930$ & $-0.914$ & $-0.980$ & $-0.948$ & $-0.890$ & $-0.763$ \\
$p_{65}$    & $\mathbf{-0.982}$ & $-0.972$ & $-0.927$ & $-0.918$ & $-0.981$ & $-0.950$ & $-0.892$ & $-0.778$ \\
$p_{70}$    & $-0.982$ & $-0.972$ & $-0.922$ & $-0.921$ & $\mathbf{-0.982}$ & $-0.952$ & $-0.905$ & $-0.810$ \\
$p_{75}$    & $-0.979$ & $-0.973$ & $-0.898$ & $-0.904$ & $-0.982$ & $-0.958$ & $-0.910$ & $-0.842$ \\
$p_{80}$    & $-0.966$ & $-0.940$ & $-0.787$ & $-0.821$ & $-0.972$ & $-0.933$ & $-0.854$ & $-0.846$ \\
$p_{85}$    & $-0.910$ & $-0.814$ & $-0.453$ & $-0.511$ & $-0.920$ & $-0.839$ & $-0.598$ & $-0.692$ \\
$p_{90}$    & $-0.722$ & $-0.628$ & $-0.024$ & $-0.193$ & $-0.736$ & $-0.664$ & $-0.215$ & $-0.480$ \\
$p_{95}$    & $-0.316$ & $-0.341$ & $+0.247$ & $+0.104$ & $-0.339$ & $-0.371$ & $+0.045$ & $-0.215$ \\
\bottomrule
\end{tabular}
\end{table}
\vfill
\newpage

\end{document}